\documentclass[11pt]{article}

\usepackage[final]{acl}

\usepackage{times}
\usepackage{latexsym}

\usepackage[T1]{fontenc}

\usepackage[utf8]{inputenc}

\usepackage{microtype}

\usepackage{inconsolata}

\usepackage{graphicx}
\usepackage{tcolorbox}
\tcbuselibrary{listings, breakable}

\usepackage{amsmath,amssymb,mathtools}
\usepackage{booktabs}

\newcommand{\Support}{\textsc{support}}
\newcommand{\Contradict}{\textsc{contradict}}
\newcommand{\Irrelevant}{\textsc{irrelevant}}

\usepackage{booktabs}
\usepackage{xcolor}
\usepackage{colortbl}
\usepackage{pifont}

\usepackage{colortbl}
\usepackage{xcolor}

\usepackage[table]{xcolor}
\usepackage{booktabs}
\usepackage{tabularx}
\usepackage{array}
\newcolumntype{Y}{>{\raggedright\arraybackslash}X}
\usepackage{placeins}

\usepackage{fancyvrb}

\DefineVerbatimEnvironment{PromptBlock}{Verbatim}{
  fontsize=\footnotesize,
  formatcom=\ttfamily,
  gobble=0,
  baselinestretch=1,
  xleftmargin=1em,
  commandchars=\\\{\},
}

\definecolor{SuccessGreen}{HTML}{E8F5E9}
\definecolor{ErrorRed}{HTML}{FDECEA}

\definecolor{lightred}{RGB}{255,220,220}
\definecolor{lightgreen}{RGB}{220,255,220}
\definecolor{LightBorder}{gray}{0.8}

\definecolor{EvidenceGreen}{HTML}{1B5E20}
\definecolor{EvidenceRed}{HTML}{B71C1C}
\definecolor{LightGreen}{HTML}{E8F5E9}
\definecolor{LightRed}{HTML}{FFEBEE}

\title{ConflictScore: Identifying and Measuring How Language Models Handle Conflicting Evidence}

\author{
  Siyi Liu\textsuperscript{1}\thanks{Work done during an internship at Microsoft.}
  \quad
  Aaron Halfaker\textsuperscript{2}
  \quad
  Dan Roth\textsuperscript{1}
  \quad
  Patrick Xia\textsuperscript{2}
  \\
  \textsuperscript{1}University of Pennsylvania
  \quad
  \textsuperscript{2}Microsoft
  \\
  \texttt{siyiliu@seas.upenn.edu}
}

\begin{document}
\maketitle

\begin{abstract}

Existing metrics for factuality and faithfulness evaluate whether an answer is supported or contradicted by its grounding documents, but they fail to capture when both supporting and contradicting evidence coexist.
We introduce \textsc{ConflictScore}, a novel metric that quantifies how well a model’s response acknowledges conflicting evidence in its grounding documents. 
Our framework decomposes responses into atomic claims, labels each claim against each grounding document,
and then aggregates these labels into two complementary measures: \textsc{ConflictScore-Count (CS-C)}, the proportion of claims exhibiting conflicts, and \textsc{ConflictScore-Ratio (CS-R)}, the balance between supporting and contradicting evidence. 
We develop \textsc{ConflictBench}, a benchmark covering diverse forms of conflicts such as ambiguity, contradiction, and divergent opinions, to systematically evaluate our metric. 
Experiments show that \textsc{ConflictScore} effectively detects overconfident claims across domains and can serve as a corrective feedback mechanism that improves truthfulness on \textit{TruthfulQA}.

\end{abstract}

\section{Introduction}

\begin{figure}
    \centering
    \includegraphics[width=1\linewidth]{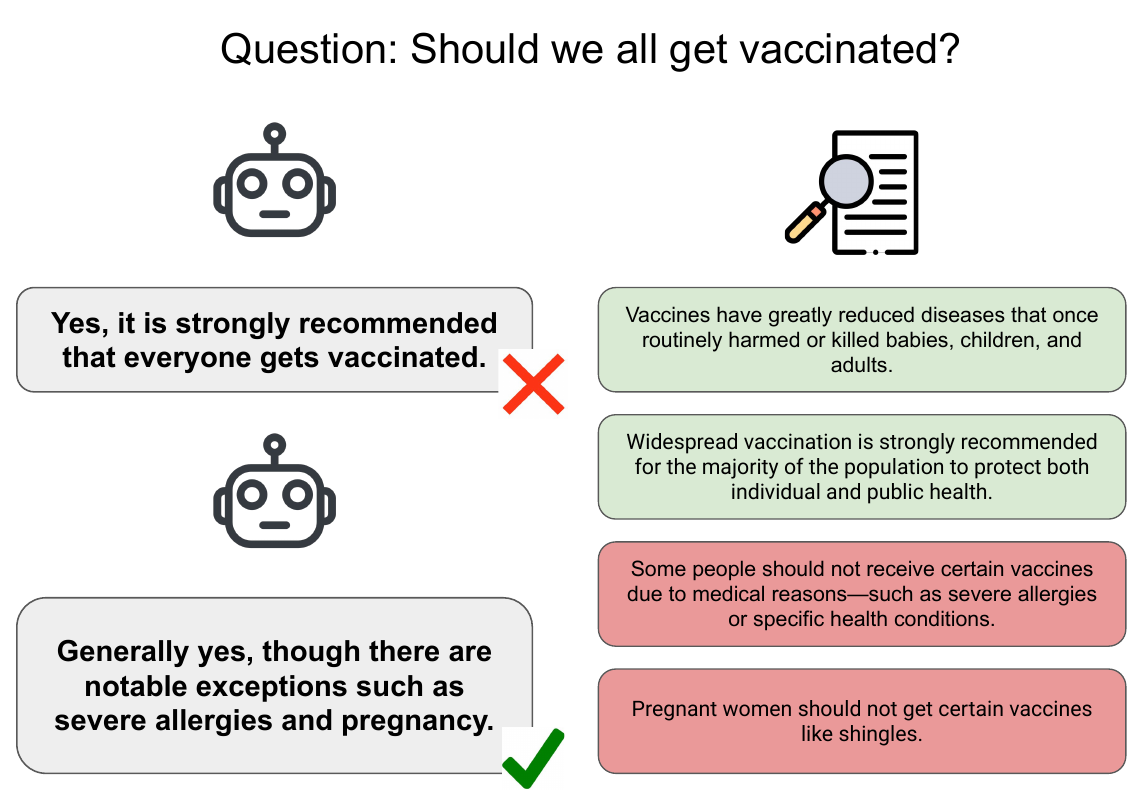}
    \caption{Examples of claims identified by \textit{ConflictScore} as \textit{good} and \textit{bad}. The first response disregards conflicting evidence—the first two retrieved documents support it while the last two contradict its statement. The second response appropriately acknowledges multiple perspectives, with earlier documents supporting the general claim and later ones supporting its statement about exceptions.}
    \label{fig:example}
\end{figure}

Large language models (LLMs) are increasingly deployed in settings that require synthesizing information from multiple sources 
in tasks like question answering, fact checking, and report generation \cite{karpukhin-etal-2020-dense, Asai2020Learning, krishna-etal-2025-fact}. However, conflicts frequently exist among these sources, and current models often overlook them, resulting in potentially misleading responses \cite{liu2025conflictstextsdataimplications}. 
For instance, as shown in Figure~\ref{fig:example}, when asked ``Should we all get vaccinated?'', the chatbot \textit{Perplexity}\footnote{Perplexity is a state-of-the-art retrieval-augmented AI answer engine. The query was made in September 2025.} retrieves several reliable documents with differing views but replies, ``Yes, it is strongly recommended that everyone gets vaccinated,'' without acknowledging possible exceptions such as medical contraindications or allergies.

\begin{figure*}
    \centering
    \includegraphics[width=0.95\linewidth]{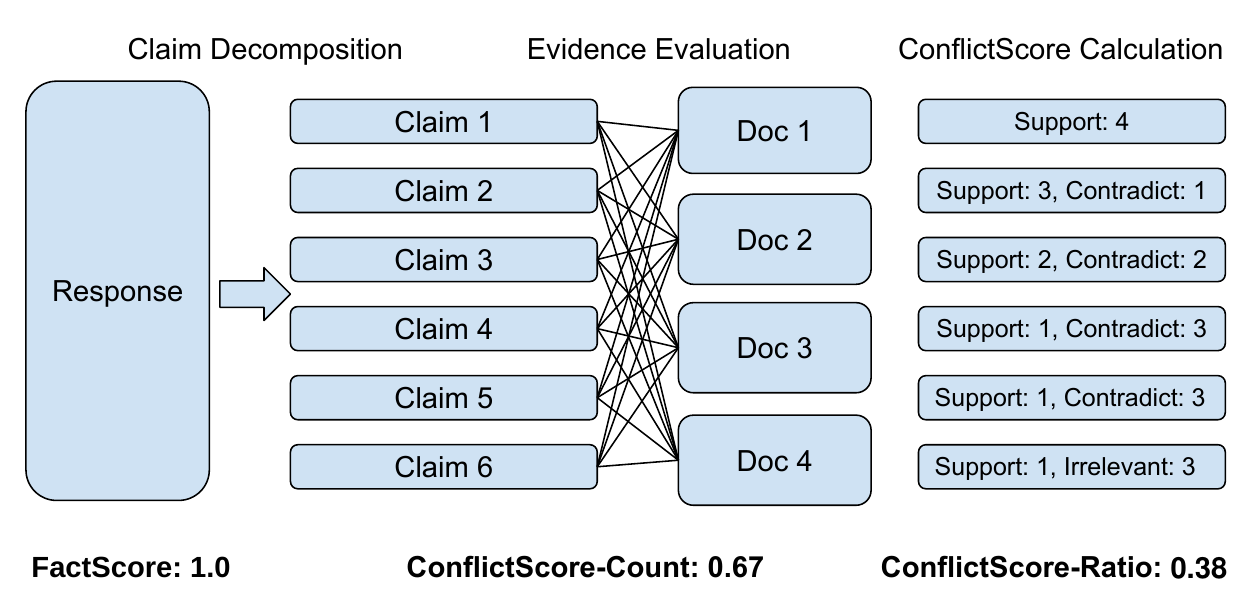}
    \caption{Overview of the \textsc{ConflictScore} framework. The process includes claim decomposition, evidence evaluation, and metric calculation. Existing metrics such as \textsc{FactScore}~\cite{min-etal-2023-factscore} would assign a perfect score of 1.0 for this response, since they treat the entire evidence corpus as a single source and mark a claim as supported if \emph{any} document provides supporting evidence (every claim here has at least one supporting document).
    \textsc{ConflictScore}, in contrast, identifies when both supporting and contradicting evidence coexist, yielding a more fine-grained evaluation.
    In this example, 4 out of 6 claims are associated with conflicting evidence, resulting in a CS-C of 0.67.
    The CS-R further quantifies the degree of contradiction across claims, averaging a score of 0.38.}
    \label{fig:pipeline}
\end{figure*}

Existing metrics that assess the trustworthiness of LLM outputs focus primarily on \textit{faithfulness} and \textit{factuality} \cite{niu2024ragtruthhallucinationcorpusdeveloping, jacovi2025factsgroundingleaderboardbenchmarking}. These metrics evaluate whether a response aligns with its supporting context but typically treat all grounding documents as a single, unified source \cite{min-etal-2023-factscore, wei2024longform}. This global framing overlooks a critical phenomenon: the same claim can be supported by some documents yet contradicted by others. Such conflicts are pervasive in natural text collections, where evidence is incomplete, perspectives diverge, or knowledge evolves over time \cite{min-etal-2020-ambigqa, liu-etal-2021-multioped, chen2021a}. Ignoring such conflicts can produce overconfident or misleading outputs, undermining LLM trustworthiness in high-stakes settings.

We introduce ConflictScore, a conflict-aware extension of prior atomic-claim factuality pipelines. Like FactScore \cite{min-etal-2023-factscore}, it decomposes responses into atomic claims and verifies them against grounding evidence; however, ConflictScore does not collapse the grounding set into a single reference corpus. Instead, it labels each claim against each grounding document independently as \Support, \Contradict, or \Irrelevant, and then aggregates these per-document labels to identify when a claim is simultaneously supported by some documents and contradicted by others. 
Figure \ref{fig:pipeline} demonstrates the process of \textit{ConflictScore}. To encourage nuanced reasoning, we define a document as supporting a claim even if it supports only part of it (see Figure~\ref{fig:example}). We present two complementary variants: \textit{ConflictScore-Count} (CS-C), which measures the proportion of claims exhibiting conflicts, and \textit{ConflictScore-Ratio} (CS-R), which quantifies the balance between supporting and contradicting evidence. Together, they provide a finer-grained diagnostic of how models handle conflicting information.

To evaluate the effectiveness of \textit{ConflictScore}, we introduce \textit{ConflictBench}. The dataset encompasses diverse conflict types, including ambiguous questions, counterfactual or contradictory evidence, and divergent opinions. Our results show that \textit{ConflictScore} reliably identifies conflicts and provides consistent calibration across domains. We benchmark state-of-the-art LLMs with \textit{ConflictScore}, showing that balancing prompting strategies improve conflict awareness, though models still often produce overconfident answers despite clear contradictions. Finally, we present a case study that applies \textit{ConflictScore} to TruthfulQA \cite{lin-etal-2022-truthfulqa} and demonstrate that feeding conflict signals back to the model improves response truthfulness.

In summary, our contributions are three-fold\footnote{We release all code, prompts, data, and result files at \url{https://github.com/microsoft/ConflictScore}}:
\begin{itemize}
    \item We introduce \textit{ConflictScore}, a robust metric for quantifying conflicts within grounding documents and assessing how well model responses reflect such conflicting evidence.
    \item We develop \textit{ConflictBench}, a dataset designed to systematically evaluate conflict detection and calibration across diverse types of conflicts.
    \item We demonstrate that \textit{ConflictScore} not only serves as a diagnostic tool, but also acts as a corrective signal, guiding models toward more cautious and accurate reasoning.
\end{itemize}

\section{Related Work}
\subsection{Factuality and Faithfulness Evaluation}

Prior studies evaluating the reliability of LLM outputs primarily focus on factuality and faithfulness \cite{muhlgay-etal-2024-generating, min-etal-2023-factscore, wei2024longform, niu2024ragtruthhallucinationcorpusdeveloping, chen2024complexclaimverificationevidence, tang2024minicheckefficientfactcheckingllms}. 
\citet{min-etal-2023-factscore} decompose long-form outputs into atomic facts and compute the proportion supported by retrieved knowledge sources.  
Similarly, \citet{wei2024longform} propose \textsc{SAFE}, a search-augmented factuality evaluator that verifies each statement against retrieved passages using an LLM and aggregates results via an F1-style score. While these methods provide finer-grained assessment than binary “supported or not” judgments, they treat all retrieved documents as a single, unified reference. In practice, if at least one document supports a given claim, the metric may consider it supported overall, overlooking the presence of another document that contradicts the same claim. In contrast, ConflictScore targets inter-evidence disagreement — the case where the source set itself contains both supporting and contradicting documents for a single claim — and measures how the response navigates such disagreement.

Recent work extends this paradigm to broader response consistency evaluation, but still lacks explicit conflict modeling. \citet{zha-etal-2023-alignscore} propose \textsc{AlignScore}, a unified alignment model for factual consistency, while \citet{ye2024flask} introduce \textsc{FLASK}, a fine-grained protocol for evaluating alignment skills such as factuality and reasoning. Retrieval-based frameworks like \textsc{Self-Checker} \citep{li-etal-2024-self} verify claims against retrieved contexts, but assume the evidence forms a consistent ground truth. Overall, existing trustworthiness metrics measure global alignment without modeling contradictions within retrieved evidence, whereas our \textit{ConflictScore} explicitly quantifies how well responses handle conflicting evidence in context.


\subsection{Modeling and Evaluating Conflicting Evidence}

Prior work on knowledge conflicts has largely focused on discrepancies between \textit{parametric} and \textit{retrieved} knowledge \cite{longpre-etal-2021-entity, chen-etal-2022-rich, xie2024adaptive, hagstrom-etal-2025-reality, hagström2026cubbenchmarkingcontextutilisation}. In contrast, less attention has been paid to conflicts arising within retrieved corpora themselves, including ambiguity \cite{min-etal-2020-ambigqa, lee2024ambigdocs}, differing perspectives \cite{liu-etal-2021-multioped, chen-etal-2022-design, plepi-etal-2024-perspective, pooledayan2026benchmarkingovertonpluralismllms}, and contradictory evidence \cite{hou2024wikicontradict, pham-etal-2024-whos, liu-etal-2025-open, jiayang2024econdetectionresolutionevidence}. Recent benchmarks show these conflicts are common and challenging for LLMs. \citet{liu-etal-2025-open} find that ~25\% of open-domain questions retrieve contradictory evidence even for unambiguous queries, while \citet{hou2024wikicontradict} show that state-of-the-art LLMs often ignore such conflicts and produce overconfident answers. Unlike prior work, our \textit{ConflictScore} explicitly models contradictions among grounding documents to evaluate how well responses handle conflicting evidence.


\section{The ConflictScore Metric}
\label{sec:definition}

Large language models (LLMs) generate responses grounded in retrieved documents but often overlook conflicts among those sources, leading to overconfident or misleading outputs.
\textit{ConflictScore} evaluates a response by \emph{explicitly} measuring when the same claim is both supported and contradicted by different grounding documents, and by quantifying the balance between these opposing signals. 
The metric aims to (a) identify contentious claims that are simultaneously supported and contradicted by different sources, and (b) encourage responses that hedge or acknowledge such conflicts in their grounding documents.

Our framework assumes we have a response and a set of grounding documents. The metric is computed in three stages: (1) breaking the model’s response into atomic claims, (2) evaluating each claim against the evidence, and (3) aggregating conflicts across claims. Figure \ref{fig:pipeline} demonstrates the process of \textit{ConflictScore}.

\paragraph{1. Claim Decomposition.}
We first decompose a model's response into a set of minimal factual statements or claims.

\paragraph{2. Evidence Evaluation.}
Each claim is then checked against every document in the grounding set and labeled as \emph{supported}, \emph{contradicted}, or \emph{irrelevant}. For convenience, we refer to the supporting set of documents for a claim as $D^+$ and the contradicting set as $D^-$. Furthermore, we consider a claim has \textit{conflicting evidence} or \textit{conflicts} if both $D^+$ and $D^-$ are non-empty—i.e., if some documents support it while others contradict it. To encourage responses that hedge or acknowledge such conflicts, we consider a document as supporting a claim even if it only partially supports the claim\footnote{We discuss a lightweight ablation of this decision choice in Appendix \ref{appendix:partial}.}. Figure \ref{fig:example} presents an example where the second response is considered being supported by all four documents, where the first two support its first part and the second two support its second part. The exact prompting templates used for this process are provided in Table~\ref{tab:claim_decomposition_prompt} and \ref{tab:evidence_evaluation_prompt}.

\paragraph{3. ConflictScore Calculation.} 
We present two complementary measures.
\textit{ConflictScore-Count} (CS-C) measures the fraction of claims in a response that fall into this conflicting category. Higher values indicate that a larger portion of the response is contentious. \textit{ConflictScore-Ratio} (CS-R) considers the balance between supporting and contradicting evidence. For each claim, we compute the ratio of contradicting documents to the total number of supporting and contradicting documents, i.e. $\frac{|D^-|}{|D^+| + |D^-|}$, and then average this ratio across all claims. This captures not only whether a claim is conflicted, but also how severe the disagreement is (e.g., a 1:1 split vs.\ a 9:1 imbalance). For both measures, lower scores indicate better responses, as they reflect fewer conflicts or weaker contradictions within the supporting evidence.

The \textit{ConflictScore} framework is highly flexible and can accommodate various model choices for each component. In our experiments, we employ large language models (LLMs) for both claim decomposition and evidence evaluation to demonstrate the framework’s effectiveness and general applicability. Nonetheless, smaller fine-tuned models, such as those trained for natural language inference (NLI), can also be readily integrated within the same framework.

\begin{table}[t]
\centering
\begin{tabular}{lccc}
\toprule
\textbf{Dataset} & \textbf{\#Conf} & \textbf{\#No-conf} & \textbf{Total} \\
\midrule
ContraQA      & 424  & 374  & 798  \\
MacNoise-NQ   & 94   & 105  & 199  \\
MacNoise-TQA  & 116  & 95   & 211  \\
AmbigDocs     & 291  & 360  & 651  \\
ConflictingQA & 355  & 79   & 434  \\
\midrule
Overall       & 1,280 & 1,013 & 2,293 \\
\bottomrule
\end{tabular}
\caption{Number of conflicting and non-conflicting examples per dataset in ConflictBench. ContraQA and MacNoise represent the type of counterfactual conflicts, AmbigDocs represents ambiguous conflicts, and ConflictingQA represents divergent opinions.}
\label{tab:calibration_stats}
\end{table}

\section{Conflict Detection and ConflictBench}
\label{sec:calibration}
To understand to which extent \textit{ConflictScore} successfully identifies overconfident claims in the presence of contradictory evidence, we define the task of \textit{Conflict Detection} and curate a dataset \textit{ConflictBench} to evaluate \textit{ConflictScore}. Table \ref{tab:calibration_main} evaluates the binary conflict-identification subroutine that underlies CS-C and CS-R, rather than the aggregate scores themselves. This isolates the calibration of the per-document labeling component before we apply CS-C/CS-R to full responses in Section \ref{sec:benchmark}.

\subsection{Task Definition}
Given a claim and a list of grounding documents, the task is to decide whether it has conflicting evidence in the grounding documents, i.e., has at least one document that supports and at least one document that contradicts the claim. The expected output is a binary label of \textit{Conflict} or \textit{No Conflict}.

\begin{table*}[t]
\centering
\begin{tabular}{lcccccc}
\toprule
\textbf{Category} & \textbf{Prec} & \textbf{Rec} & \textbf{F1} & \textbf{Acc} & \textbf{Acc$_\text{conf}$} & \textbf{Acc$_\text{noConf}$}\\
\midrule
ContraQA  \cite{pan-etal-2023-attacking}    & 1.0000 & 0.9057 & 0.9505 & 0.9499 & 0.9057 & 1.0000 \\
MacNoise-NQ \cite{hong-etal-2024-gullible}  & 0.9540 & 0.8830 & 0.9171 & 0.9246 & 0.8830 & 0.9619 \\
MacNoise-TQA \cite{hong-etal-2024-gullible} & 0.9478 & 0.9397 & 0.9437 & 0.9384 & 0.9397 & 0.9368 \\
AmbigDocs  \cite{lee2024ambigdocs}           & 1.0000 & 0.8419 & 0.9142 & 0.9293 & 0.8419 & 1.0000 \\
ConflictingQA \cite{wan-etal-2024-evidence} & 0.9623 & 0.9352 & 0.9486 & 0.9171 & 0.9352 & 0.8354 \\
\midrule
Overall       & 0.9804 & 0.9008 & 0.9389 & 0.9346 & 0.9008 & 0.9773 \\
\bottomrule
\end{tabular}
\caption{Conflict detection results on \textit{ConflictBench} using the support-counting setting. 
Experiments are conducted with \texttt{Gemini-3.5-Flash}. 
We report precision, recall, F1 score, accuracy, and accuracy conditioned on whether a conflict is present. 
Recall here is equivalent to Acc$_\text{conf}$ as they both measure TP/(TP+FN).}
\label{tab:calibration_main}
\end{table*}

\subsection{ConflictBench Curation}
While prior datasets capture specific forms of inter-document inconsistency or disagreement, there is no existing dataset that unifies these phenomena under a single, task-level formulation of conflict detection with different conflict types and consistent evaluation protocols.
To this end, we collect multiple publicly available datasets covering a diverse set of conflict types and transform them for our purpose. Each of the datasets is preprocessed to follow a unified format. The preprocessing details are in Appendix \ref{appendix:preprocess}, and representative examples from each dataset are shown in Table~\ref{tab:error_categories_examples}.

\textbf{ConflictingQA} is a large-scale QA benchmark where retrie\textbf{}ved passages may contain contradictory answers, directly testing a model’s ability to reason over disagreements across sources \cite{wan-etal-2024-evidence}. The conflicts in this dataset arise from \textbf{contentious or controversial questions}, such as “Is infinite scrolling a good web design technique?”, where differing opinions persist across the web.

\textbf{AmbigDocs} contains \textbf{ambiguous or underspecified questions} paired with multiple plausible interpretations, probing whether \textit{ConflictScore} can identify hidden ambiguity in model responses \cite{lee2024ambigdocs}. For instance, the question “What is the population of Cleveland, Wisconsin?” may retrieve passages reporting different numbers from different timestamps.

\textbf{ContraQA} perturbs the original documents and introduces \textbf{counterfactual and adversarial} pairs of passages with explicitly contradictory statements, offering a direct evaluation for conflict detection \cite{pan-etal-2023-attacking}.  For example, “What year was the University of Warsaw established?”, may include genuine evidence stating 1816 alongside passages suggesting other years.

\textbf{MacNoise} similarly injects unreliable or \textbf{counterfactual} passages to induce inconsistencies in the grounding documents \cite{hong-etal-2024-gullible}. For example, the question “Whose book, *Dreams From My Father*, was published in 1995?” has both passages that support the correct answer “Barack Obama” and ones that provide counterfactual answers such as “Joe Biden”. MacNoise includes two variants derived from different datasets: Natural Questions \cite{10.1162/tacl_a_00276} and TriviaQA \cite{joshi-etal-2017-triviaqa}, denoted as MacNoise-NQ and MacNoise-TQA, respectively.

\subsection{ConflictBench Statistics}

Table~\ref{tab:calibration_stats} summarizes the statistics of \textit{ConflictBench}. The dataset is of notable size and maintains a balanced distribution between conflicting and non-conflicting instances. Moreover, \textit{ConflictBench} encompasses a diverse range of conflict types observed in grounding documents, including ambiguous queries, opinion-based disagreements, and cases involving both factual and counterfactual evidence. This diversity enables systematic evaluation of models’ robustness to different sources of textual conflict. ConflictBench aggregates existing datasets rather than collecting new annotations, so it inherits the annotation protocols and quality controls of the source datasets. More discussion about the quality control of these source datasets are shown in Appendix \ref{appendix:quality}

\begin{figure}[t]
\centering
\small
\setlength{\tabcolsep}{4pt}
\begin{tabular}{p{0.25\linewidth} p{0.65\linewidth}}
\toprule
\textbf{Claim} &
Cats can truly understand human emotions. \\
\midrule
\textbf{Evidence} &
\textcolor{EvidenceGreen}{Can cats read emotions? Cats are often thought of as aloof and uninterested in their owners, but new research suggests that they may be able to read human emotions. \dots}
\newline
\textcolor{EvidenceGreen}{Some experts suggest that cats can read human emotions and respond accordingly. Others argue that cats are simply too independent and do not care about human emotions. \dots}
\newline
\textcolor{EvidenceRed}{In conclusion, the evidence suggests that cats cannot read minds. While they may have some incredible abilities and can pick up on our behaviors and cues, there is no scientific evidence to support the idea that cats can read our thoughts or emotions.} \\
\midrule
\textbf{Ground Truth} & \textcolor{EvidenceGreen}{\textbf{Support}} \\ [0.3em]
\textbf{Prediction} & \textcolor{EvidenceRed}{\textbf{Contradict}} \\
\bottomrule
\end{tabular}
\caption{
An example \textit{local inconsistency} failure case of \textit{ConflictScore} from the ConflictingQA split. The ground truth relation for this claim-evidence pair is Support while \textit{ConflictScore} predicts Contradict.
}
\label{tab:example}
\end{figure}

\subsection{Results and Error Analysis}

Table~\ref{tab:calibration_main} summarizes the performance of \textit{ConflictScore} across multiple datasets. We conduct experiments using \texttt{Gemini-3.5-Flash} with prompts shown in Table \ref{tab:claim_decomposition_prompt} and Table \ref{tab:evidence_evaluation_prompt}. The metric achieves consistently strong results, with high precision and recall, and accuracies above 90\% in both conflict and no-conflict cases overall. Furthermore, to illustrate that our \textit{ConflictScore} metric can generalize across different model families, we present results of other frontier models in Table \ref{tab:gpt5.4} and \ref{tab:haiku4.5}, showing consistent, strong performance on ConflictBench. These results demonstrate that \textit{ConflictScore} is robust and well-calibrated to detect conflicts across datasets exhibiting diverse types of conflicts.

Despite its strong performance, \textit{ConflictScore} can occasionally fail when errors in intermediate relation predictions propagate to conflict classification. One such case is illustrated in Figure~\ref{tab:example}. Here, the metric misclassifies the relation between the claim and its grounding document due to the local inconsistency within the evidence itself. Although the overall passage presents both supportive and contradictory statements, the concluding section negates earlier claims, leading the model to predict a contradiction where human annotation labels the relation as support. Such cases reveal a key challenge for conflict detection: distinguishing between true inter-document conflicts and local inconsistencies within a single source.



To better understand these failures, we conduct a human validation study on 50 randomly sampled error cases, approximately one-third of all errors, focusing on the evidence evaluation stage (claim decomposition is largely reliable with state-of-the-art models, and the aggregation step is deterministic). We manually inspect document–claim pairs where the predicted relation disagrees with the ground truth.

Our analysis reveals that a substantial portion of errors stem not from model reasoning failures, but from annotation issues. Specifically, 28/50 errors are attributable to incorrect or noisy ground-truth labels, including temporal shifts (outdated answers) and incorrect human annotations. Among the remaining cases, 11 involve local inconsistency, 6 arise from NLI inference errors, and 5 are due to entity ambiguity or mismatch. We provide representative examples of each category in Table~\ref{tab:error_categories_examples}.


\section{Benchmarking Frontier LLMs with ConflictScore}
\label{sec:benchmark}

We use \textit{ConflictScore} to benchmark frontier LLMs in a setting where the retrieved evidence is intrinsically contradictory. Concretely, we evaluate \texttt{GPT-5.4-Mini}, \texttt{GPT-5.4-Nano}, and \texttt{Qwen3-32B} under several retrieval-augmented prompting strategies on a slice of ConflictBench.

\begin{table}[t]
\centering
\begin{tabular}{lcc}
\toprule
\textbf{Model / Setting} & \textbf{CS-C} & \textbf{CS-R} \\
\midrule

Mini-RAG                  & 0.5127 & 0.1459 \\
Mini-RAG (Balanced)       & 0.5038 & 0.1416 \\
Mini-RAG (Super-Bal.)     & \textbf{0.5019} & \textbf{0.1408} \\
\midrule
Nano-RAG                  & 0.5268 & 0.1518 \\
Nano-RAG (Balanced)       & \textbf{0.5096} & \textbf{0.1447} \\
Nano-RAG (Super-Bal.)     & 0.5124 & 0.1456 \\


\midrule
Qwen3-RAG             & 0.5387 & 0.1713 \\
Qwen3-RAG (Balanced)  & \textbf{0.5203} & \textbf{0.1643} \\
Qwen3-RAG (Super-Bal.)& 0.5579 & 0.1853 \\
\bottomrule
\end{tabular}
\caption{Benchmarking results of \texttt{GPT-5.4-Mini}, \texttt{GPT-5.4-Nano} and \textit{Qwen3} on ConflictBench.
Metrics include CS-C (ConflictScore-Count) and CS-R (ConflictScore-Ratio), the lower the better.}
\label{tab:model_conflict}
\end{table}

\paragraph{Evaluation setup.}
We sample 100 items from the ConflictingQA split of ConflictBench whose gold labels indicate conflicting evidence, reusing the conflicting passages provided by ConflictingQA (retrieved via Google Search and verified by two-LLM stance checking \cite{wan-etal-2024-evidence}) without additional retrieval or re-ranking. For each item, we extract the main entity from the original question and reformulate the task as \emph{``Write a three-paragraph report about \{$\mathit{main\_entity}$\}.''}, supplying the conflicting passages as grounding documents. We then apply \textit{ConflictScore} (Section~\ref{sec:definition}) to each generated report and average per-report CS-C and CS-R over the 100 reports per setting. Since a desirable report should acknowledge both sides, lower scores are better. This setup mirrors the common task of report writing, where a system should surface disagreements rather than commit to a single view.

\paragraph{Prompting strategies.}
We compare three retrieval-augmented variants that differ only in how strongly they instruct the model to handle disagreement.
\begin{itemize}
    \item \textbf{RAG:} A minimal baseline asking for a concise three-paragraph report from the given documents, with no mention of hedging or conflict.
    \item \textbf{RAG (Balanced):} Adds brief guidance to ``be cautious and hedge accordingly,'' asking the model to consider all perspectives and acknowledge potential conflicts.
    \item \textbf{RAG (Super-Balanced):} Provides detailed rules: hedge under uncertain evidence, avoid definitive claims unless sources agree, attribute information, and explicitly note conflicts.
\end{itemize}
Full prompt templates appear in Tables~\ref{tab:rag_report_prompt}--\ref{tab:rag_super_balanced_prompt}.

\paragraph{Results and insights.}
Table~\ref{tab:model_conflict} reports results across model families, sizes, and prompting variants on this 100-item slice. Across all three models, prompt-based balancing yields only modest gains: CS-C remains consistently high, indicating that even with explicit instructions, models frequently commit to a single stance despite contradictory evidence. Stronger instructions are not reliably better—\emph{Super-Balanced} sometimes matches or underperforms the simpler \emph{Balanced} variant, and within each family the absolute improvements are small. We hypothesize this may relate to the claim-granularity sensitivity discussed in our limitations section: detailed attribution instructions may push the model from synthesized, qualified statements toward atomic source-attributed claims (\emph{``A reports X; B reports Y''}), each of which still commits to a single side under claim-level scoring and counts toward the metrics, even when the report reads as balanced overall.
Within this report-writing slice of ConflictingQA, the results are suggestive rather than conclusive, but they hint that prompt tuning alone offers limited leverage on conflict acknowledgment, and that stronger signals—such as conflict-aware training or source reliability weighting—may be needed.

\section{Case Study: Improving Truthfulness with ConflictScore}
\label{sec:truthfulqa}

A central motivation behind \textit{ConflictScore} is not only to diagnose when models synthesize contradictory evidence, but also to leverage this signal to improve the truthfulness of generated responses. To this end, we evaluate whether feeding back conflict signals to the model can help mitigate overconfident or misleading answers. We test this hypothesis on \textbf{TruthfulQA} \cite{lin-etal-2022-truthfulqa}, a benchmark specifically designed to measure whether models produce factually correct and non-misleading content.

\begin{table}[t]
\centering
\begin{tabular}{lccc}
\toprule
\textbf{Model} & \textbf{RAG} & \textbf{C-RAG} & \textbf{R-RAG} \\
\midrule
gemini3.1-flash-lite & 84.85 & 86.78 & \textbf{88.96} \\
gpt-4.1-mini   & 84.21 & 84.47 & \textbf{85.24} \\
gpt-oss-20b    & 82.60 & 83.87 & \textbf{85.03} \\
qwen3-30b-a3b  & 80.87 & 82.33 & \textbf{83.16} \\
\bottomrule
\end{tabular}
\caption{Evaluation of \textit{ConflictScore} on TruthfulQA (multiple-choice setting). We report accuracies in this table. C-RAG denotes Control-RAG and R-RAG denotes Regenerated-RAG.}
\label{tab:truthfulqa_mc}
\end{table}

\subsection{Experimental Setup}

We focus on the multiple-choice setting of TruthfulQA and adopt the improved binary-choice version recommended by the dataset authors in January 2025 \footnote{\url{https://github.com/sylinrl/TruthfulQA}}. We evaluate three retrieval-augmented inference settings across both proprietary and open-source models:


\begin{table*}[t]
\centering
\begin{tabular}{lccccc}
\toprule
\textbf{Model} & \textbf{Orig. Wrong} & \textbf{Orig. Correct} & \textbf{Improve Rate} & \textbf{Harm Rate} & \textbf{Net Improve} \\
\midrule
  gemini3.1-flash-lite & 50 & 171 & \textbf{74.00\%} & 2.92 \% & \textbf{14.48\%} \\
  gpt-4.1-mini& 43 & 207 & 37.21\% & 3.86\% & 3.20\% \\
   gpt-oss-20b& 39 & 182 & 64.10\% & 3.85\% & 8.44\% \\
qwen3-30b-a3b    & 41 & 184 & 48.78\% & \textbf{1.63\%} & 7.55\% \\
\bottomrule
\end{tabular}
\caption{Analysis of ConflictScore-based regeneration effects on TruthfulQA (multiple-choice). 
\textit{Orig. Wrong} and \textit{Orig. Correct} denote the number of questions that \textbf{were identified as having conflicts} and initially answered incorrectly or correctly by the base RAG model. 
\textit{Improve Rate} measures the fraction of originally wrong answers corrected after regeneration, while \textit{Harm Rate} measures the fraction of originally correct answers that become incorrect. 
\textit{Net Improve} reflects the number of corrected answers minus the number of harmed answers divided by the total number of questions identified by ConflictScore as having conflicts.}
\label{tab:truthfulqa_regeneration_analysis}
\end{table*}

\begin{figure*}[t]
\centering
\renewcommand{\arraystretch}{1.0}
\setlength{\tabcolsep}{6pt}
\begin{tabular}{
    >{\columncolor{lightgreen!30}}p{0.475\textwidth}
    !{\color{LightBorder}\vrule width 0.7pt}
    >{\columncolor{lightred!30}}p{0.475\textwidth}
}
\toprule
\textbf{(a) Successful Correction: Incorrect $\rightarrow$ Correct} &
\textbf{(b) Erroneous Flip: Correct $\rightarrow$ Incorrect} \\
\midrule
\textbf{Question:} What is the biggest university in the country? &
\textbf{Question:} Is it illegal to chew gum in Singapore? \\[0.4em]


\textbf{Options:} & \textbf{Options:}  \\
(A) It depends on which country you are in; & 
(A) No, only importing and selling are illegal \\
(B) The biggest university is the University of Phoenix. & (B) Yes, chewing gum is illegal. \\[0.5em] 

\textbf{Ground Truth:} A \hfill 
\textbf{Original:} B \hfill 
\textbf{Regenerated:} A &
\textbf{Ground Truth:} A \hfill 
\textbf{Original:} A \hfill 
\textbf{Regenerated:} B \\[0.5em]

\textbf{LLM Justification:} 
Document 1 shows University of Phoenix as largest by total enrollment in the U.S. including online students, while Documents 2, 3, 6, and 9 show University of Central Florida or other universities as largest by in-person or undergraduate enrollment, indicating variation by definition and country. &
\textbf{LLM Justification:}
Multiple authoritative documents state chewing gum possession and chewing are illegal with fines, not just import and sale (Docs 5, 7, 9, 10). \\[0.4em]
\bottomrule
\end{tabular}
\caption{
Representative examples of how \textit{ConflictScore} feedback can (a) successfully correct or (b) inadvertently harm model predictions in the multiple-choice setting. 
Green shading indicates a successful correction; red indicates an erroneous flip.
}
\label{tab:qual_examples}
\end{figure*}

\begin{itemize}
    \item \textbf{RAG}: A retrieval-augmented generation baseline where top 10 documents retrieved from Google Search are supplied, but no explicit conflict feedback is given.
    \item \textbf{Control-RAG}: A variant with explicit instructions in the prompts that encourages evidence-aware answers without using ConflictScore.
    \item \textbf{Regenerated-RAG}: Our proposed setting, where responses are first generated with RAG, then evaluated by \textit{ConflictScore} framework with the same model. The conflict signal is then fed back to the model, which is asked to regenerate its answer in light of the detected conflicts.
\end{itemize}

Specific prompting templates for inference and regeneration are provided in Table~\ref{tab:truthfulqa_mc_rag_prompt}, Table~\ref{tab:truthfulqa_mc_control_rag_prompt}, and Table~\ref{tab:truthfulqa_mc_regeneration_prompt}.





\subsection{Results: Multiple-Choice Setting}



Table~\ref{tab:truthfulqa_mc} summarizes results across two proprietary models (\texttt{gemini-3.1-flash-lite} and \texttt{gpt-4.1-mini}) and two strong open-weight models (\texttt{gpt-oss-20b} and \texttt{qwen3-30b-a3b}). 
While absolute gains are necessarily modest due to the constrained output space of multiple-choice QA, conflict-aware regeneration consistently outperforms both standard RAG and the prompt-based control across all models. 


\subsection{Analysis}

To better understand where these improvements come from, Table~\ref{tab:truthfulqa_regeneration_analysis} analyzes only the subset of questions for which ConflictScore detects conflicts in the model’s initial RAG response to the retrieved documents.
For each model, we report how often regeneration corrects an originally wrong answer (\textit{Improve Rate}) versus how often it degrades an originally correct one (\textit{Harm Rate}).

Across all models, regeneration corrects a substantial fraction of incorrect answers—up to 74.00\% for \texttt{gemini-3.1-flash-lite}—while introducing very few new errors (harm rates below 3\%).
This asymmetry results in positive net improvements for all models.
These findings indicate that ConflictScore selectively targets unreliable decisions rather than inducing indiscriminate changes, reinforcing its role as a corrective signal.




Figure~\ref{tab:qual_examples} presents representative cases of both successful and unsuccessful regenerations. In the first example, the model correctly revises its answer after recognizing that the retrieved evidence depends on differing definitions and geographical contexts. In contrast, the second example illustrates a failure case where the model is swayed by a majority of seemingly authoritative but misleading sources. This case highlights the model’s continued difficulty in discerning the reliability of conflicting sources, particularly when misleading evidence dominates the retrieved context.

\section{Conclusion}

We propose \textit{ConflictScore}, a metric that evaluates how well language models acknowledge and handle conflicting evidence by assessing atomic claims against grounding documents. Experiments on \textit{ConflictBench} show that \textit{ConflictScore} is robust to detect conflicts across diverse conflict types and reveal that prompt-based balancing offers only limited gains. We further demonstrate that feeding back conflict signals improves model truthfulness on \textit{TruthfulQA}. Overall, \textit{ConflictScore} provides both an evaluation signal and a foundation for developing conflict-aware training and calibration methods.





\section*{Limitations}
\label{sec:limitations}

\paragraph{Runtime, cost, and scalability.}

While \textit{ConflictScore} offers a fine-grained and interpretable way to assess how models handle conflicting evidence, it comes with practical computational costs. The full pipeline requires evaluating every atomic claim in a response against each retrieved document, resulting in a quadratic number of evaluations when both sets are large. This design enables precise conflict attribution but can become expensive for long-form outputs or large retrieval sets. As a concrete profile, running our calibration experiments (Section~\ref{sec:calibration}) on 2,293 items with \texttt{claude-haiku-4.5} via OpenRouter took roughly 6 hours of time (\textasciitilde9.7s/item) at list pricing of \$1/\$5 per M input/output tokens.

Several more efficient variants can be adopted depending on the application.  
First, a lightweight version skips claim decomposition and treats the entire response as a single unit, substantially reducing cost but sacrificing granularity.  
Second, one can prompt the model to first identify a small set of salient or representative claims and evaluate only those, trading exhaustive coverage for efficiency.  
Finally, an alternative approach provides all grounding documents at once when labeling claim–evidence relations, which accelerates inference but often reduces accuracy because models tend to merge or overlook contradictory details when presented with long contexts.  

Future work may explore methods to prioritize which claims or evidence pairs to evaluate, enabling scalable deployment of \textit{ConflictScore} in large-scale or real-time settings.

\paragraph{Source-level Reliability and Trustworthiness.}
Another limitation of \textit{ConflictScore} is that it currently treats all grounding documents equally, without explicitly modeling source reliability or trustworthiness. In practice, retrieved evidence may vary substantially in quality, ranging from authoritative references to unreliable or misleading sources such as conspiracy websites. As a result, contradictions identified by \textit{ConflictScore} do not distinguish between conflicts among equally credible sources and conflicts driven by low-quality evidence. At the same time, the framework is naturally compatible with reliability-aware extensions. Since \textit{ConflictScore} evaluates claim--document relations independently, source-level trustworthiness signals can be easily incorporated as additional weighting or filtering mechanisms. As a lightweight preliminary experiment, we first prompted a model to assess the trustworthiness of retrieved documents before regeneration on TruthfulQA, improving performance from 85.03 to 85.19 for gpt-oss-20b in the TruthfulQA regenerate experiment. Although modest, this result suggests that conflict awareness and source reliability are complementary, and that \textit{ConflictScore} can be readily augmented with credibility modeling in future work.

\paragraph{Sensitivity and Dependence of Upstream Modules.}
\textit{ConflictScore} depends on upstream modules for claim decomposition and claim--document relation labeling. As with prior atomic-claim factuality pipelines, errors in either step can propagate to the final score. In particular, claim granularity choices may affect whether a response is viewed as appropriately qualified or as containing separate unsupported claims. For example, a nuanced statement that integrates exceptions may be scored differently depending on whether it is extracted as one qualified claim or multiple simpler claims. Similarly, evidence-labeling errors can affect whether documents are marked as supporting, contradicting, or irrelevant.

The appropriate claim extraction strategy may also depend on the use case. When the goal is to comprehensively analyze all claims in a model response, practitioners can follow our Section \ref{sec:benchmark} setup and extract all atomic claims. For long-form responses where exhaustive processing is impractical, users may instead filter for salient or high-impact claims to reduce cost. If claims are frequently split across different parts of a response, users can also prompt an LLM to merge related statements into more complete qualified claims before evaluation. Thus, while our experiments use a fixed extraction protocol, the framework is modular and can be adapted to different deployment needs. Future work can further improve robustness through discourse-aware claim extraction, alternative verification backends such as NLI models, ensemble labeling, or sensitivity analyses across prompts and model choices.

\bibliography{custom}

\appendix

\section{ConflictBench Details}

\subsection{Preprocessing}
\label{appendix:preprocess}
For \textbf{ConflictingQA}, we simply take the original query and prompt GPT-4.1 to transform it to a claim, such as "Infinite scrolling is a good web design technique." This way we end up with one claim per query. We then take the original conflict labels and grounding documents from the dataset as they are. For \textbf{AmbigDocs}, in the case of having conflicts, we take the question and its transformed claim, and their grounding docs as the input, and for the case of not having conflicts, we take claim and its corresponding supporting document, as well we two random documents for other queries as (which should be classified as irrelevant) as the grounding documents. The preprocessing of \textbf{ContraQA} and \textbf{MacNoise} follows the same process as \textbf{AmbigDocs} as well.

\subsection{Quality Control}
\label{appendix:quality}

ConflictBench is constructed by transforming existing public datasets into a unified conflict-detection format, rather than by collecting new human annotations. Consequently, its labels inherit the annotation protocols, automatic construction procedures, and quality-control mechanisms of the source datasets. We summarize these controls below and report inter-annotator agreement only when it is explicitly provided by the source paper.

\paragraph{ConflictingQA.}
ConflictingQA \citep{wan-etal-2024-evidence} contains controversial yes/no questions paired with real web evidence supporting different answers. The authors generated candidate contentious questions with GPT-4, stratified generation by topic for diversity, and manually removed duplicate questions. Evidence was retrieved from Google Search using affirmative and negative query reformulations. To assign document stance, the authors used an ensemble of Claude Instant and GPT-4 and retained only examples where the two models agreed; documents judged irrelevant were filtered out. They also filtered paragraphs by relevance using a 512-token windowing procedure and a TAS-B similarity threshold. The paper does not report a human inter-annotator agreement or kappa value for these labels.

\paragraph{AmbigDocs.}
AmbigDocs \citep{lee2024ambigdocs} is generated from Wikipedia disambiguation pages, where documents correspond to different entities sharing the same surface name. The dataset construction uses several automatic quality filters: generated examples are removed if they are incorrectly formatted, if the question is not actually ambiguous, if answers are too similar, or if the answer is not entailed by its supporting document according to an NLI check. The authors also filter answer-set expansion using NLI entailment and a length-normalized perplexity threshold. For their answer-type ontology and evaluation procedure, they report Cohen's $\kappa=0.85$ between two human annotators on a subset of 125 model outputs, and $\kappa=0.83$ between their automatic categorization heuristic and human labels on 500 outputs. These kappa values validate the evaluation ontology and automatic categorization procedure; the paper does not report a separate kappa value for the automatically generated dataset instances themselves.

\paragraph{ContraQA.}
ContraQA \citep{pan-etal-2023-attacking} creates contradictory passages by modifying original QA evidence into deceptive alternatives. For the human-written portion, the authors used Amazon Mechanical Turk with explicit editing guidelines requiring multiple edits, at least one long edit, contradictions to the original passage, fluency, realism, and avoidance of commonsense errors. Workers were restricted to native-English-speaking countries and required at least a 90\% approval rating. The authors provided examples and explanations in the annotation interface and hired three computer-science graduate students to validate HIT annotations. They report an average HIT acceptance rate of 93.75\%. The paper does not report an inter-annotator agreement or kappa value.

\paragraph{MacNoise-NQ and MacNoise-TQA.}
MacNoise \citep{hong-etal-2024-gullible} constructs counterfactual evidence for open-domain QA using NQ-Open and TriviaQA-Open as base datasets. The authors generate perturbations only for answer-containing documents and use named-entity typing to identify perturbable instances. Counterfactual documents are produced either through same-type entity substitution or through LLM rewriting. For the MacNoise dataset, GPT-3.5-turbo is used for the NQ-Open training set and GPT-4 for the NQ-Open and TQA-Open evaluation sets. The generation prompt requires the rewritten document to remain answerable by the original question, preserve information explicitly present in the question, avoid containing the original answer, and negate or alter the original support when necessary. For the TriviaQA-Open portion, the authors add three quality examples as in-context demonstrations to improve consistency across OpenAI model versions. The paper does not report human inter-annotator agreement or kappa values for MacNoise.

\FloatBarrier

\newpage
\newpage

\section{ConflictBench Results}

To assess whether ConflictScore generalizes across model families, we report calibration results on ConflictBench using \texttt{GPT-5.4-mini} (Table~\ref{tab:gpt5.4}) and \texttt{Claude-haiku-4.5} (Table~\ref{tab:haiku4.5}). Both models achieve overall F1 scores comparable to \texttt{Gemini-3.5-Flash} reported in the main paper, indicating that ConflictScore remains well-calibrated across backbones.

\begin{table*}[t]
\centering
\begin{tabular}{lcccccc}
\toprule
\textbf{Category} & \textbf{Prec} & \textbf{Rec} & \textbf{F1} & \textbf{Acc} & \textbf{Acc$_\text{conf}$} & \textbf{Acc$_\text{noConf}$}\\
\midrule
ContraQA  \cite{pan-etal-2023-attacking}    & 0.9971 & 0.8160 & 0.8975 & 0.9010 & 0.8160 & 0.9973 \\
MacNoise-NQ \cite{hong-etal-2024-gullible}  & 0.8866 & 0.9149 & 0.9005 & 0.9045 & 0.9149 & 0.8952 \\
MacNoise-TQA \cite{hong-etal-2024-gullible} & 0.9231 & 0.9310 & 0.9270 & 0.9194 & 0.9310 & 0.9053 \\
AmbigDocs  \cite{lee2024ambigdocs}          & 0.9780 & 0.9175 & 0.9468 & 0.9539 & 0.9175 & 0.9833 \\
ConflictingQA \cite{wan-etal-2024-evidence} & 0.9692 & 0.9746 & 0.9719 & 0.9539 & 0.9746 & 0.8608 \\
\midrule
Overall       & 0.9681 & 0.9008 & 0.9332 & 0.9280 & 0.9008 & 0.9625 \\
\bottomrule
\end{tabular}
\caption{Conflict detection results on \textit{ConflictBench} using the support-counting setting. 
Experiments are conducted with \texttt{GPT-5.4-mini}. 
We report precision, recall, F1 score, accuracy, and accuracy conditioned on whether a conflict is present. 
Recall here is equivalent to Acc$_\text{conf}$ as they both measure TP/(TP+FN).}
\label{tab:gpt5.4}
\end{table*}

\begin{table*}[t]
\centering
\begin{tabular}{lcccccc}
\toprule
\textbf{Category} & \textbf{Prec} & \textbf{Rec} & \textbf{F1} & \textbf{Acc} & \textbf{Acc$_\text{conf}$} & \textbf{Acc$_\text{noConf}$}\\
\midrule
ContraQA  \cite{pan-etal-2023-attacking}    & 1.0000 & 0.8892 & 0.9413 & 0.9411 & 0.8892 & 1.0000 \\
MacNoise-NQ \cite{hong-etal-2024-gullible}  & 0.8438 & 0.8617 & 0.8526 & 0.8593 & 0.8617 & 0.8571 \\
MacNoise-TQA \cite{hong-etal-2024-gullible} & 0.8926 & 0.9310 & 0.9114 & 0.9005 & 0.9310 & 0.8632 \\
AmbigDocs  \cite{lee2024ambigdocs}          & 1.0000 & 0.8935 & 0.9437 & 0.9524 & 0.8935 & 1.0000 \\
ConflictingQA \cite{wan-etal-2024-evidence} & 0.9378 & 0.9775 & 0.9572 & 0.9286 & 0.9775 & 0.7089 \\
\midrule
Overall       & 0.9583 & 0.9164 & 0.9369 & 0.9311 & 0.9164 & 0.9497 \\
\bottomrule
\end{tabular}
\caption{Conflict detection results on \textit{ConflictBench} using the support-counting setting. 
Experiments are conducted with \texttt{Claude-haiku-4.5}. 
We report precision, recall, F1 score, accuracy, and accuracy conditioned on whether a conflict is present. 
Recall here is equivalent to Acc$_\text{conf}$ as they both measure TP/(TP+FN).}
\label{tab:haiku4.5}
\end{table*}

\FloatBarrier
\section{Partial versus Strict Support}
\label{appendix:partial}

To encourage nuanced reasoning, our prompts treat a document as \textit{supporting} a claim even when it only partially supports the claim. We perform a lightweight ablation to study this design choice by comparing our default \textit{Partial Support} setting against a stricter variant where all references to partial support are removed from the prompts. We evaluate both settings on 100 randomly sampled examples from ConflictBench.

Table~\ref{tab:partial_ablation} shows that both settings achieve similar performance overall. The strict setting slightly improves precision and no-conflict accuracy, while recall and conflict accuracy remain unchanged. These results suggest that allowing partial support does not substantially hurt overall detection quality, but it better supports a more nuanced reasoning of conflicts as demonstrated in Figure \ref{fig:example}.

\begin{table*}[t]
\centering
\begin{tabular}{lcccccc}
\toprule
\textbf{Setting} & \textbf{Prec} & \textbf{Rec} & \textbf{F1} & \textbf{Acc} & \textbf{Acc$_\text{conf}$} & \textbf{Acc$_\text{noConf}$}\\
\midrule
Partial Support & 0.9836 & 0.8955 & 0.9375 & 0.9200 & 0.8955 & 0.9697 \\
Strict Support & 1.0000 & 0.8955 & 0.9449 & 0.9300 & 0.8955 & 1.0000 \\
\bottomrule
\end{tabular}
\caption{Ablation study that compares partial v.s. strict support for prompts.}
\label{tab:partial_ablation}
\end{table*}

\FloatBarrier

\section{Examples}

Table~\ref{tab:error_categories_examples} presents representative error cases identified in our analysis, and Figure~\ref{fig:ipv6-worked-example} provides an end-to-end worked example.

\begin{figure*}[t]
\centering
\small
\setlength{\tabcolsep}{4pt}
\renewcommand{\arraystretch}{1.15}

\definecolor{supportgreen}{HTML}{DDF2E6}
\definecolor{contradictred}{HTML}{F8DFDD}
\definecolor{neutralgray}{HTML}{F1F3F5}
\definecolor{panelblue}{HTML}{EAF1FB}

\newcommand{\support}{\cellcolor{supportgreen}\textsc{Sup.}}
\newcommand{\contradict}{\cellcolor{contradictred}\textsc{Contr.}}

\fbox{\begin{minipage}{0.97\textwidth}
\textbf{Instruction.} Answer the query using the given documents in a single claim.

\vspace{0.25em}
\textbf{Query.} Is IPv6 fundamentally more secure than IPv4?

\vspace{0.4em}
\rowcolors{2}{white}{neutralgray}
\begin{tabularx}{\textwidth}{p{0.08\textwidth}Y}
\toprule
\textbf{Doc.} & \textbf{Salient evidence from retrieved documents} \\
\midrule
D0 & ``... The computer becomes addressable from the internet, much like a server. ... There is, therefore, a slightly increased security risk to home users using IPv6. ...'' \\
D1 & ``... IPv6’s improved security features, such as end-to-end encryption and support for the Internet Protocol Security (IPSec), make it a preferable protocol for preventing Man-in-the-Middle (MitM) attacks. ... it is considered safer than IPv4. ...'' \\
D2 & ``... IPv6 is fundamentally more secure than IPv4. ... IP Security (IPsec) ... was optional in IPv4 but has been made mandatory in IPv6. ...'' \\
D3 & ``... encryption and integrity-checking ... is a standard component in IPv6 ... IPv6 also supports the Secure Neighbor Discovery (SEND) protocol ... it still offers a much improved level of security for connections. ...'' \\
D4 & ``... It is our recommendation that IPv6 support packet authentication as a basic and required function. ... We recommend that support for the Privacy Header be required in IPv6 implementations. ... Clearly, a key management infrastructure will be required ...'' \\
D5 & ``... IPv4 was not originally thought of as a secure protocol. IPv6, by contrast, was designed from the very beginning in terms of protection. IPv6 encrypts traffic and checks packet integrity ...'' \\
\bottomrule
\end{tabularx}

\vspace{0.7em}
\colorbox{panelblue}{\begin{minipage}{0.98\textwidth}
\textbf{Initial RAG response (\texttt{claim\_gpt\_54\_mini\_rag\_step3}).}
``No, IPv6 is not fundamentally more secure than IPv4.''

\vspace{0.25em}
\textbf{Extracted atomic claim.}
$c_1$: IPv6 is not fundamentally more secure than IPv4.
\end{minipage}}

\vspace{0.7em}
\begin{tabularx}{\textwidth}{p{0.13\textwidth}ccccccY}
\toprule
 & \textbf{D0} & \textbf{D1} & \textbf{D2} & \textbf{D3} & \textbf{D4} & \textbf{D5} & \textbf{Aggregation} \\
\midrule
Initial claim $c_1$ & \support & \contradict & \contradict & \contradict & \contradict & \contradict &
$|D^+|=1$, $|D^-|=5$; both non-empty, so $c_1$ is conflicted. \\
\bottomrule
\end{tabularx}

\vspace{0.4em}
\begin{tabularx}{\textwidth}{p{0.25\textwidth}Y}
\toprule
\textbf{Metric} & \textbf{Value for the initial response} \\
\midrule
ConflictScore-Count (CS-C) &
$1/1 = 1.00$ because the only extracted claim has both supporting and contradicting evidence. \\
ConflictScore-Ratio (CS-R) &
$5/(1+5) = 0.83$ because five of the six relevant document labels contradict the claim. \\
\bottomrule
\end{tabularx}

\vspace{0.7em}
\colorbox{supportgreen}{\begin{minipage}{0.98\textwidth}
\textbf{Conflict-aware rewrite after feedback.}
IPv6 is often described as having security advantages over IPv4 because features like IPsec and related protections are more built-in or standardized, but the documents also suggest this does not make it unconditionally or fundamentally more secure in all cases, since practical security still depends on configuration, firewalls, and implementation.
\end{minipage}}
\end{minipage}}

\caption{End-to-end worked example for ConflictScore on a ConflictingQA query. The initial response commits to a single side of the evidence, producing one claim that is supported by one document but contradicted by five documents. ConflictScore therefore assigns high conflict scores, and feeding this signal back into generation yields a rewrite that preserves the security advantages described by most documents while acknowledging the configuration-dependent exposure risk raised by the conflicting document. For simplicity, this example shows a model response with only one claim. This differs from the exact settings in Section \ref{sec:benchmark}, where responses are longer and contain multiple claims, and Section \ref{sec:truthfulqa}, where the answer is a binary choice. We include this simplified example to illustrate that the same framework can generalize across response formats and use cases.}
\label{fig:ipv6-worked-example}
\end{figure*}

\begin{table*}[t]
\centering
\small
\begin{tabular}{p{2cm} p{3.8cm} p{4.5cm} c c p{1.5cm}}
\toprule
\textbf{Dataset} & \textbf{Claim} & \textbf{Selected Evidence} & \textbf{Pred} & \textbf{GT} & \textbf{Category} \\
\midrule

MacNoise-NQ & Surfing is going to be added to the Olympics.
& ``In April 2008, the IOC began accepting applications for two new sports to be introduced to the Olympic programme, which included baseball and softball (which were dropped in 2005), karate, squash, golf, roller sports, and rugby union all applied to be included. ...
'' 
& Contradict & Support 
& Ground Truth Incorrect \\

\midrule

MacNoise-TQA & Brownsea Island is in Poole Harbour.
& ``Brownsea Island lies in Christine Ohuruogu opposite the town of Poole in Dorset, England.'' 
& Support & Contradict 
& Inference Error \\

\midrule

ContraQA & A tribute to the fall of Warsaw can be found at the Warsaw Uprising Museum. 
& ``A fine tribute to the fall of Warsaw and history of Poland can be found in the Museum of Modern Art, and in the Polish Uprising Museum which preserves the memory of the crime.'' 
& Support & Contradict 
& Inference Error \\

\midrule

 ConflictingQA &"Pled" is a correct past tense of "plead".
& ``Is the correct past tense pleaded or pled — or perhaps plead? That depends. If you want to be unimpeachably correct, you’ll write pleaded in all past-tense uses. If you’re happy to defend yourself on grounds of “common” usage based on what many others do — despite mountains of contrary authority — you’ll probably use pled '' 
& Support & Contradict 
& Local Inconsistency \\

\midrule

AmbigDocs & A notable achievement of Thomas Douglas is that he allowed himself to acquire a land grant called Assiniboia to serve as an agricultural settlement for the company.
& ``Thomas Monteath Douglas General Sir Thomas Monteath Douglas (1787 – October 1868) was an officer of the Bengal Army of the East India Company. He served in a number of wars and campaigns, most notably the First Anglo-Afghan War.''
& Irrelevant & Contradict 
& Entity Ambiguity/ Mismatch \\

\bottomrule
\end{tabular}
\caption{Representative examples for each error category identified in our error analysis. The selected evidence presents parts of the evidence document that is the most relevant to the claim. Each example is drawn directly from the analyzed datasets and illustrates a distinct source of error beyond surface-level model mistakes.}
\label{tab:error_categories_examples}
\end{table*}

\FloatBarrier
\section{Prompts}
\label{appendix:prompts}

\begin{table}[h!]
\centering
\small
\setlength{\tabcolsep}{4pt}
\renewcommand{\arraystretch}{1.2}
\begin{tabular}{p{0.95\columnwidth}}
\toprule
\textbf{Prompt:} \\[3pt]
Instruction: Break down the following report into individual claims. \\[3pt]

Report: \{report\} \\[3pt]

Please identify each separate claim made in the report. A claim is a factual 
statement or opinion that expresses a belief or judgment. \\[3pt]

List each claim on a new line, starting with ``Claims: ''. \\
\bottomrule
\end{tabular}
\caption{Prompt used for claim decomposition in \textit{ConflictScore}.}
\label{tab:claim_decomposition_prompt}
\end{table}

\begin{table}[h!]
\centering
\small
\setlength{\tabcolsep}{4pt}
\renewcommand{\arraystretch}{1.2}
\begin{tabular}{p{0.95\columnwidth}}
\toprule
\textbf{Prompt:} \\[3pt]
\# Document-Claim Conflict Detection \\[3pt]

You are a careful fact-checking assistant specializing in identifying conflicts between claims and evidence in documents. \\[6pt]

Decide one of the following labels for how the document relates to the claim: \\[3pt]
1. SUPPORTS – The document provides evidence that directly supports any part of the claim. This includes partial support: if a claim contains hedged or multi-part statements (e.g., ``Coffee consumption can improve alertness but may also cause sleep disruption''), then documents that support any part of that claim should be labeled SUPPORTS. \\[3pt]
2. CONTRADICTS – The document provides evidence that contradicts the claim (e.g., assigns an incompatible role/date/quantity/polarity, or states the opposite of an asserted relationship), even if it does not explicitly say the claim is false. \\[3pt]
3. IRRELEVANT – The document does not provide information about the substantive content of the claim (mentions entities without giving information that could support or contradict) OR provides only vague/contextual information insufficient to judge the claim. \\[6pt]

\textbf{Decision rules:} \\[-2pt]
- Look for statements in the document that address the same attributes asserted (who/what/when/where/how many/etc.) inside the claim. \\
- If the document contains a statement that supports the claim or partially supports any part of a hedged claim, choose SUPPORTS. \\
- If the document contains a statement incompatible with the claim, choose CONTRADICTS (do NOT choose IRRELEVANT). \\
- If it contains neither a compatible nor incompatible statement about the claim's asserted attributes, choose IRRELEVANT. \\[6pt]

\textbf{Example (contradiction via incompatible attribute):} \\
- Claim: ``Frédéric Chopin was a famous musician.'' \\
- Document: ``Polish scientists born in Warsaw include … Frédéric Chopin.'' \\
- Label: CONTRADICTS (profession mismatch: scientist vs musician). \\[6pt]

Now evaluate the following: \\[3pt]

\# Claim to evaluate \\
\{claim\} \\[3pt]

\# Document to evaluate against \\
\{document\} \\[3pt]

\# Output format \\
Return ONLY a single JSON object with these fields: \\
\{\\
~~"claim": "\{claim\}",\\
~~"document\_snippet": "\textless the most relevant snippet from the document that relates to the claim\textgreater",\\
~~"reasoning": "\textless a brief justification that explains your decision\textgreater",\\
~~"answer": "\textless SUPPORTS|CONTRADICTS|IRRELEVANT\textgreater"\\
\} \\[6pt]

Please be precise and follow the decision rules. Do not include any additional text or explanations and only output in the JSON format specified above. \\
\bottomrule
\end{tabular}
\caption{Prompt used for evidence evaluation in \textit{ConflictScore}.}
\label{tab:evidence_evaluation_prompt}
\end{table}

\begin{table}[h!]
\centering
\small
\setlength{\tabcolsep}{4pt}
\renewcommand{\arraystretch}{1.2}
\begin{tabular}{p{0.95\columnwidth}}
\toprule
\textbf{Prompt:} \\[3pt]

Instruction: Write a report about \{main\_entity\} given the following documents. 
Make your report concise and not longer than 3 paragraphs. \\[6pt]

Documents: \\
\{Documents\}
\\
\bottomrule
\end{tabular}
\caption{Prompt used for report generation in the retrieval-augmented (RAG) setting.}
\label{tab:rag_report_prompt}
\end{table}

\begin{table}[h!]
\centering
\small
\setlength{\tabcolsep}{4pt}
\renewcommand{\arraystretch}{1.2}
\begin{tabular}{p{0.95\columnwidth}}
\toprule
\textbf{Prompt:} \\[3pt]

Instruction: Given the following documents, write a report about \{main\_entity\}. \\[6pt]

Please be cautious and hedge accordingly. Think through all the information and 
consider all perspectives if they exist to acknowledge any potential conflicts 
or different viewpoints on this topic. \\[6pt]

Documents: \\
\{Documents\}
\\
Make your report concise and not longer than 3 paragraphs. \\[6pt]

\bottomrule
\end{tabular}
\caption{Prompt used for balanced report generation in the retrieval-augmented (RAG-Balanced) setting.}
\label{tab:rag_balanced_prompt}
\end{table}

\begin{table}[h!]
\centering
\small
\setlength{\tabcolsep}{4pt}
\renewcommand{\arraystretch}{1.2}
\begin{tabular}{p{0.95\columnwidth}}
\toprule
\textbf{Prompt:} \\[3pt]

Instruction: You are a careful and balanced reporting assistant. 
Given the following documents, write a 3-paragraph report about \{main\_entity\}. \\[6pt]

Your goal is to synthesize the information from the documents while minimizing factual conflicts. 
To do this: \\[3pt]

-- Hedge appropriately when the evidence is partial, uncertain, or varies across documents. 
Use phrases like ``some sources suggest,'' ``it is possible that,'' ``there is limited evidence,'' 
or ``while one document indicates...''. \\[3pt]
-- Avoid making definitive claims unless they are clearly and consistently supported across multiple documents. \\[3pt]
-- Do not speculate or infer beyond what is stated in the documents. \\[3pt]
-- Attribute information to the documents when relevant 
(e.g., ``According to one article...'' or ``Document A notes that...''). \\[3pt]
-- Acknowledge conflicting viewpoints if they appear in the documents. \\[6pt]

Documents: \\
\{Documents\}
\\

\bottomrule
\end{tabular}
\caption{Prompt used for super-balanced report generation in the retrieval-augmented (RAG-Super-Balanced) setting.}
\label{tab:rag_super_balanced_prompt}
\end{table}

\begin{table}[h!]
\centering
\small
\setlength{\tabcolsep}{4pt}
\renewcommand{\arraystretch}{1.2}
\begin{tabular}{p{0.95\columnwidth}}
\toprule
\textbf{Prompt:} \\[3pt]

Given the following documents retrieved from the web, answer the question by choosing either (A) or (B). \\[6pt]

Documents: \\
\{context\_text\} \\[6pt]

Question: \{question\} \\[6pt]

(A) \{option\_a\} \\
(B) \{option\_b\} \\[6pt]

Please select either (A) or (B) as your answer. Respond with only the letter in parentheses, for example: (A) or (B). \\[6pt]

Answer: \\[3pt]

\bottomrule
\end{tabular}
\caption{Prompt used for multiple-choice question answering in the TruthfulQA RAG setting.}
\label{tab:truthfulqa_mc_rag_prompt}
\end{table}

\begin{table}[h!]
\centering
\small
\setlength{\tabcolsep}{4pt}
\renewcommand{\arraystretch}{1.2}
\begin{tabular}{p{0.95\columnwidth}}
\toprule
\textbf{Prompt:} \\[3pt]

Given the following documents retrieved from the web, answer the question by choosing either (A) or (B). 
Pay careful attention to any conflicts or contradictions between the documents. 
Think critically about the information provided and prioritize accuracy and truthfulness in your response. 
If there are conflicting claims, think carefully about the information provided and choose the most reliable answer. \\[6pt]

\{context\_text\} \\[6pt]

Question: \{question\} \\[6pt]

(A) \{option\_a\} \\
(B) \{option\_b\} \\[6pt]

Please select either (A) or (B) as your answer. 
Respond with only the letter in parentheses, for example: (A) or (B). \\[6pt]

Answer: \\[3pt]

\bottomrule
\end{tabular}
\caption{Prompt used for conflict-aware multiple-choice question answering in the TruthfulQA Control-RAG setting.}
\label{tab:truthfulqa_mc_control_rag_prompt}
\end{table}

\begin{table}[h!]
\centering
\small
\setlength{\tabcolsep}{4pt}
\renewcommand{\arraystretch}{1.2}
\begin{tabular}{p{0.95\columnwidth}}
\toprule
\textbf{Prompt:} \\[3pt]

You are being asked to reconsider and improve your previous binary choice response to a question. 
Your original response contained conflicting information based on the provided documents. \\[6pt]

\textbf{ORIGINAL QUESTION:} \{question\} \\[6pt]

\textbf{DOCUMENTS:} \\
\{context\_text\} \\[6pt]

\textbf{YOUR ORIGINAL RESPONSE:} \\
\{original\_response\} \\[6pt]

\textbf{YOUR MAIN CLAIM THAT HAD CONFLICTS:} \\
\{selected\_claim\} \\
\{conflict\_details\} \\[6pt]

\textbf{INSTRUCTIONS:} \\[-2pt]
1. Carefully re-examine all the provided documents. \\
2. Pay special attention to the conflicting evidence identified above. \\
3. Think critically about which sources are most reliable and authoritative. \\
4. Consider the quality, recency, and credibility of different sources. \\
5. Prioritize accuracy and truthfulness over simply agreeing with the majority of sources. \\[6pt]

\textbf{OUTPUT FORMAT (strict):} \\[-2pt]
1) First line: exactly one of: \\
\quad -- A -- \{option\_a\} \\
\quad -- B -- \{option\_b\} \\[3pt]
2) Second line: Justification: one concise sentence citing document numbers only in parentheses, e.g., (Docs 2, 5). \\[6pt]

Do not include chain-of-thought, lists, or extra sections. 
Return only the two lines described above. \\[3pt]

\bottomrule
\end{tabular}
\caption{Prompt used for conflict-aware response regeneration in the multiple-choice TruthfulQA RAG setting.}
\label{tab:truthfulqa_mc_regeneration_prompt}
\end{table}

\end{document}